\pdfoutput=1

\documentclass[11pt]{article}

\usepackage{EACL2023}

\usepackage{times}
\usepackage{latexsym}

\usepackage[T1]{fontenc}

\usepackage[utf8]{inputenc}

\usepackage{microtype}

\usepackage{inconsolata}

\usepackage{times}
\usepackage{latexsym}
\usepackage{soul}
\usepackage{amsmath}
\usepackage{booktabs}
\usepackage{bm}
\usepackage{cleveref}
\usepackage{graphicx}
\usepackage{tabularx}
\usepackage{soul}
\usepackage{xcolor}
\usepackage{todonotes}
\usepackage{multirow}
\usepackage{xpatch}
\usepackage{blindtext}
\usepackage{fdsymbol}
\usepackage{microtype}
\usepackage{hyperref}
\usepackage{adjustbox}
\usepackage{textcomp}
\usepackage{xparse}
\usepackage{enumitem}
\usepackage{makecell}
\usepackage{subcaption}
\usepackage[linesnumbered,vlined,ruled]{algorithm2e}
\SetAlCapNameFnt{\small}
\SetAlCapFnt{\small}

\crefformat{section}{\S#2#1#3} %
\crefformat{subsection}{\S#2#1#3}
\crefformat{subsubsection}{\S#2#1#3}

\newcommand{\samsum}[1]{\textsc{SAMSum}}
\newcommand{\dialogsum}[1]{\textsc{DialogSum}}
\newcommand{\mixandmatch}[1]{\textsc{MixAndMatch}}
\newcommand{\confit}[1]{\textsc{ConFiT}}
\newcommand{\ctrldiasumm}[1]{\textsc{CtrlDiaSumm}}
\newcommand{\cods}[1]{\textsc{CODS}}
\newcommand{\modelshort}[1]{\textsc{Swing}}
\newcommand{\pgn}[1]{\textsc{PGN}}
\newcommand{\cliff}[1]{\textsc{CLIFF}}
\newcommand{\conseq}[1]{\textsc{ConSeq}}

\definecolor{c2}{RGB}{218,0,0}

\definecolor{lightblue}{RGB}{212, 235, 255}
\definecolor{lightorange}{RGB}{255, 204, 168}
\definecolor{lightyellow}{RGB}{255, 255, 168}
\definecolor{lightred}{RGB}{255, 168, 168}
\definecolor{lightgreen}{RGB}{190, 255, 168}
\definecolor{gold}{rgb}{0.83, 0.69, 0.22}
\sethlcolor{lightblue}

\newcommand\hlc[2]{\sethlcolor{#1} \hl{#2}}
\iftrue

\NewDocumentCommand{\steeve}
{ mO{} }{\textcolor{gold}{\textsuperscript{\textit{Steeve}}\textsf{\textbf{\small[#1]}}}}
\newcommand\kathy[1]{{\color{purple!45!blue}[\textit{#1}]$_{-KM}$}}
\else
\newcommand{\steeve}[1]{}
\newcommand\kathy[1]{}
\fi
\definecolor{gold}{rgb}{0.83, 0.69, 0.22}

\title{\modelshort~\includegraphics[width=1.1em]{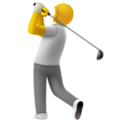}: Balancing Coverage and Faithfulness\\ for Dialogue Summarization}

\author{Kung-Hsiang Huang\textsuperscript{$\spadesuit$}\thanks{~~Work done while interning at Amazon.}~~~
Siffi Singh\textsuperscript{$\diamondsuit$}~~~ 
Xiaofei Ma\textsuperscript{$\diamondsuit$}~~~ 
Wei Xiao\textsuperscript{$\diamondsuit$}~~~ \\
    {\bfseries 
    Feng Nan\textsuperscript{$\diamondsuit$}~~~
     Nick Dingwall\textsuperscript{$\diamondsuit$}~~~
    William Yang Wang\textsuperscript{$\diamondsuit$} ~~~
    Kathleen McKeown\textsuperscript{$\diamondsuit$}} \\
  \textsuperscript{$\spadesuit$}University of Illinois Urbana-Champaign 
   ~~~
  \textsuperscript{$\diamondsuit$}AWS AI Labs \\
  \texttt{khhuang3@illinois.edu} \\
  \texttt{\{siffis, xiaofeim, weixiaow, nanfen, nickding, wyw, mckeownk\}@amazon.com} \\
  }

\begin{document}
\maketitle

\begin{abstract}
Missing information is a common issue of dialogue summarization where some information in the reference summaries is not covered in the generated summaries.
To address this issue, we propose to utilize natural language inference (NLI) models to improve coverage while avoiding introducing factual inconsistencies. Specifically, we use NLI to compute fine-grained training signals to encourage the model to generate content in the reference summaries that have not been covered, as well as to distinguish between factually consistent and inconsistent generated sentences. Experiments on the \dialogsum~ and \samsum~ datasets confirm the effectiveness of the proposed approach in balancing coverage and faithfulness, validated with automatic metrics and human evaluations. Additionally, we compute the correlation between commonly used automatic metrics with human judgments in terms of three different dimensions regarding coverage and factual consistency to provide insight into the most suitable metric for evaluating dialogue summaries.\footnote{We release our source code for research purposes: \\\url{https://github.com/amazon-science/AWS-SWING}.}

\end{abstract}

\section{Introduction}

Dialogue summarization is a text generation task that aims to produce a compact summary given a piece of conversation. Conventional approaches to dialogue summarization rely on features of conversation data \cite{DBLP:conf/slt/GooC18, li-etal-2019-keep, oya-etal-2014-template}. Recently, the rise of large pre-trained language models (LMs) has enabled coherent and fluent summaries to be generated without these features. 
However, low coverage and factual inconsistency remain two pressing issues as studies have shown that the summaries generated from these pre-trained LMs often do not fully cover the reference \cite{liu-chen-2021-controllable, tang-et-al-2022-confit} and that the generated summaries are often not factually consistent with the inputs \cite{zhang-etal-2020-optimizing, maynez-etal-2020-faithfulness, cao-wang-2021-cliff}. 
If an unfaithful dialogue summarization model with low coverage is deployed for public use, it could spread misinformation and generate misleading content that only covers partial facts of a conversation. Hence, we are urgently in need of a solution to improve coverage without negatively impacting faithfulness for dialogue summarization.

\begin{figure}[bt]
    \centering
    \includegraphics[width=0.9\linewidth]{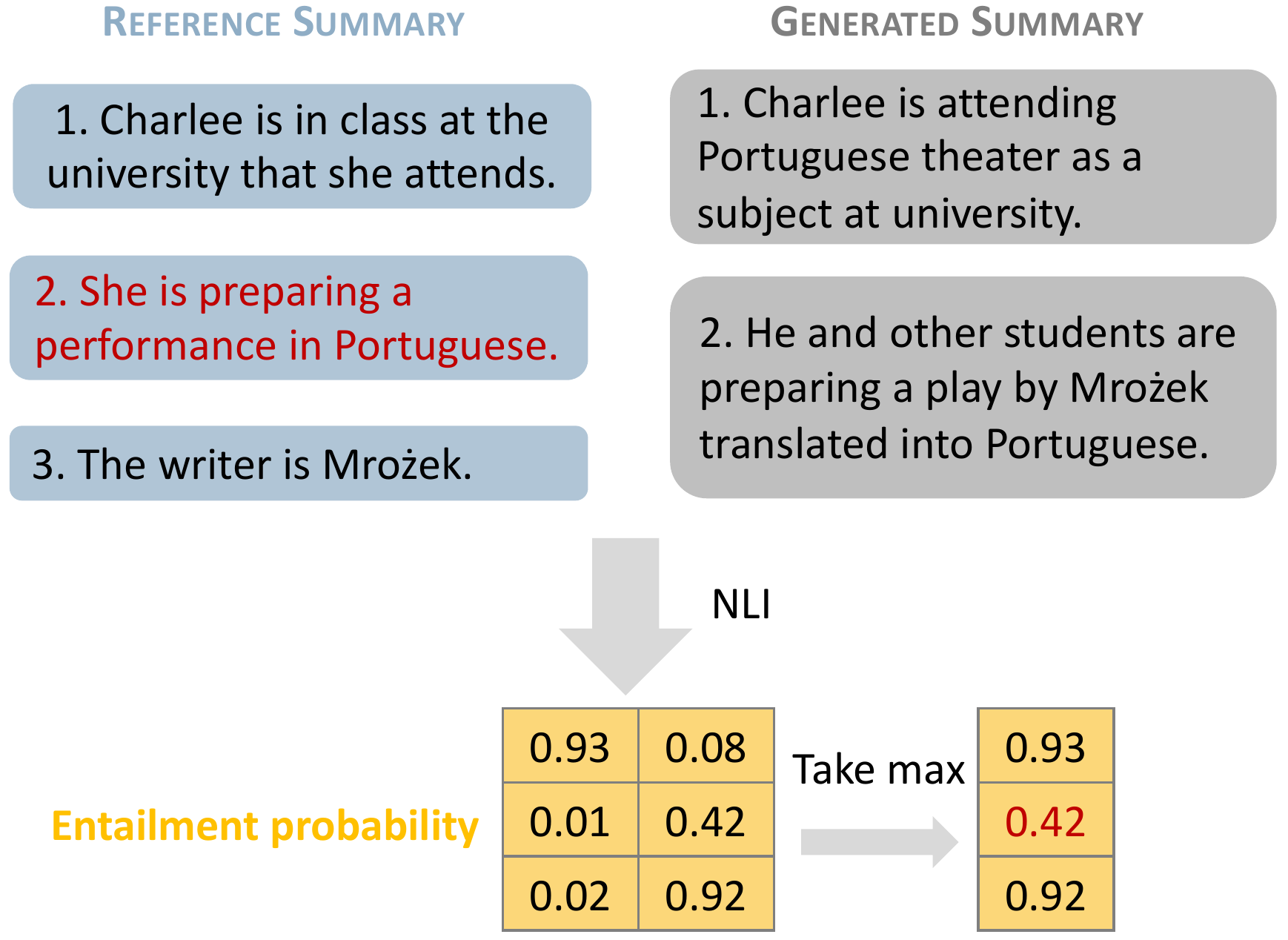}
    \caption{An illustration of how NLI can help determine whether a reference sentence is covered by the generated summary. We compute the entailment probability from each reference sentence (i.e. premise) to each generated sentence (i.e. hypothesis). By taking the max value along the row dimension, the resulting vector denotes the probability that each reference sentence entails a sentence in the generated summary. In this example, the entailment probability for the second reference sentence is low, indicating that this sentence is likely not covered by the generated summary. }
    \label{fig:toy_example}
\end{figure}

\begin{figure*}[t]
    \centering
    \includegraphics[width=0.9\linewidth]{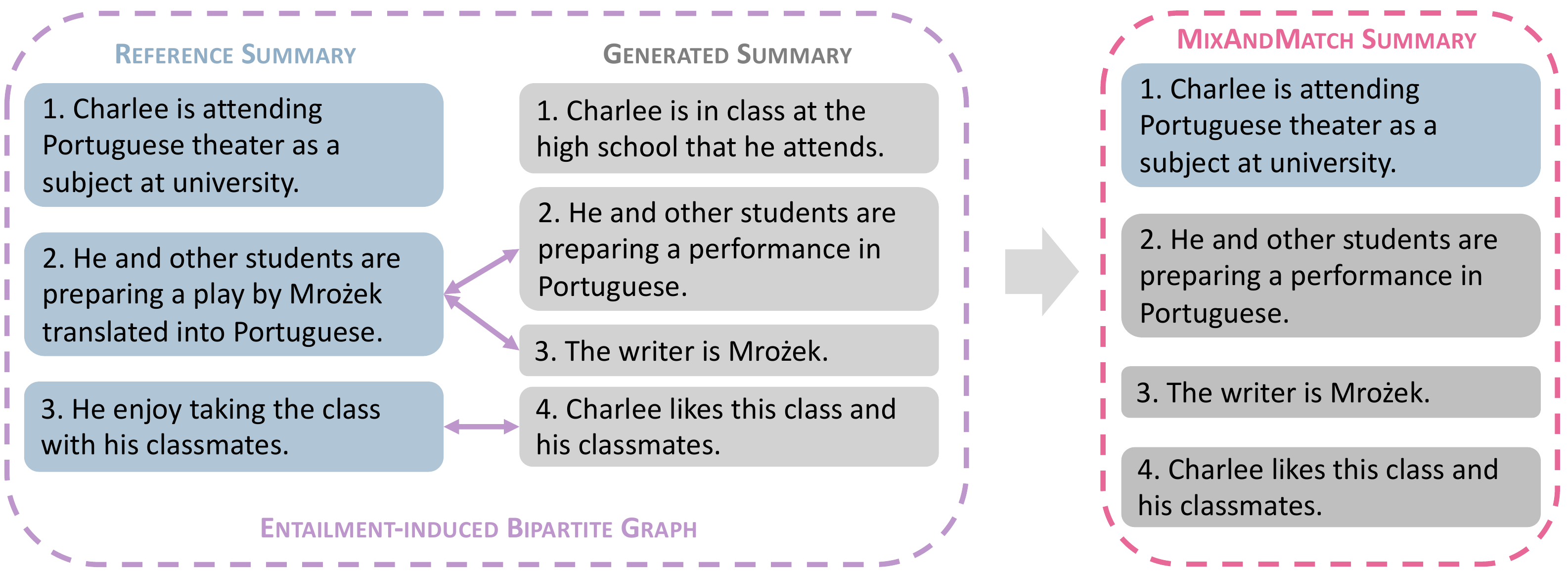}
    \caption{Illustration of how an entailment-induced bipartite graph is built and how a \mixandmatch~ summary is derived. With the NLI model, we determine which sentences from each summary contain equivalent information by computing the entailment probabilities between pairs of generated sentences and reference sentences, as indicated by the purple edges. Based on the graph, we determine that the generated summary does not cover the first reference sentence and that the first generated sentence is not faithful. Hence, the \mixandmatch~ summary is formed by combining the first reference sentence and the second to the fourth generated sentence.}
    \label{fig:mix-n-match}
\end{figure*}
Relatively little work addresses coverage and factual inconsistency for dialogue summarization. Some work addresses the issue of unfaithfulness with a controllable generation framework guided by person named entities \cite{liu-chen-2021-controllable} or summary sketches \cite{wu-etal-2021-controllable}. \citet{tang-et-al-2022-confit} categorize factual inconsistencies for dialogue summarization into different types of errors, such as missing information and wrong reference. Their framework integrates a contrastive loss and a self-supervised loss to reduce multiple types of errors. However, a great portion ($> 40\%$) of their outputs does not cover the full content of the reference summary. Thus, it is important to address coverage and factual consistency synergistically in dialogue summarization.
The issue where the content in the reference does not occur in the generated summary is known as the missing information issue \cite{liu-chen-2021-controllable, tang-et-al-2022-confit}. In this work, we aim to mitigate missing information in the summary while being faithful to the dialogue.

We propose \modelshort~\includegraphics[width=1.1em]{images/person-golfing_1f3cc-fe0f.png}, \underline{\textbf{S}}ummarizing Dialogue \underline{\textbf{Wi}}th \underline{\textbf{N}}LI \underline{\textbf{G}}uidance. Our approach samples a summary from the model and utilizes natural language inference (NLI) to determine (1) the faithfulness of each generated sentence and (2) whether each reference sentence has been covered by the generated summary. An example is shown in \Cref{fig:toy_example}. Based on the results computed by NLI, two losses are proposed to encourage the model to generate missing information and distinguish between factually consistent and inconsistent generated sentences.

Our contributions can be summarized as follows:
\begin{itemize}[noitemsep,nolistsep]
  \item We propose \modelshort~, a dialogue summarization framework that effectively addresses missing information through two losses computed using NLI. The first loss encourages the model to recover content missing from the reference summaries. The second loss instructs the model to differentiate between factually consistent and inconsistent generated sentences.\looseness=-1
  \item Our approach achieves the best performance in mitigating missing information on two public dialogue summarization datasets, \dialogsum~ \cite{chen-etal-2021-dialogsum} and \samsum~ \cite{gliwa-etal-2019-samsum}, as validated by automatic metrics and human judges.
  \item We measure the correlation of human judgments with conventional and recently developed automatic metrics to provide intuition for future research on evaluating the faithfulness and coverage of dialogue summaries.

\end{itemize}

\section{Method}

Upon analyzing the dialogue summaries in \samsum~, we observe that dialogues are often summarized linearly, consistent with the findings of \citet{wu-etal-2021-controllable}. 
Therefore, we segment the summaries into sentences and use a natural language inference (NLI) model to provide finer-grained training signals at the sentence level for two goals: (1) encourage generating sentences in the reference summaries that have not been covered by the generated sentences and (2) differentiate factually consistent generated sentences from inconsistent ones. To achieve these goals, we first determine the faithfulness of each sentence using an entailment-induced bipartite graph (\cref{sec:bipartite_graph}). Then, we propose two new losses addressing each challenge in turn: an \textbf{Uncovered Loss} that encourages the model to recover missing information (\cref{sec:uncovered_loss}) and a \textbf{Contrastive Loss} that brings closer the representations of the reference summary and the generated sentences that contain equivalent information to some sentences in the reference summary (\cref{sec:contrastive_loss}). For the rest of this paper, we use \textit{reference sentence} and \textit{generated sentence} to refer to a sentence in the reference summary and the generated summary, respectively.

\begin{algorithm}[t]
\small

\caption{Entailment-induced Bipartite Graph}
\label{alg:bipartite_graph}
\KwIn{
A reference summary $S^* = \{s^*_1, ..., s^*_n \}$, a generated summary $S = \{s_1, ..., s_m \}$;
}
\KwOut{The bipartite mapping $\phi$ between sentences in $S^*$ and $S$;}

Initialize $\phi$ as a zero matrix of size $n \times m$\, where $n = |S^*|$ and $m = |S|$\;
Let $\tau$ be the entailment threshold\;
// Resolve 1-to-many mappings\;
\For{$i \gets 1$ to $n$} {
    $V \gets \emptyset$ \;
    \For {$j \gets 1$ to $m$}{
        \If{$p_{\text{ent}}(s^*_i, s_j) > \tau$ and $\phi(i, j) = 0$}{
            $V \gets V \cup j$\;
        }
        
    }
    $s_V \gets$ Concatenate sentences in $\{s_v \forall v \in V \}$\;
    \If{$V \text{ is consecutive and}$ $p_{\text{ent}}(s_V, s^*_i) > \tau$ }{
        \For {$v \in V$}{
            $\phi(i, v) \gets 1$\;
        }
    }
}
// Resolve many-to-1 mappings\;
\For{$j \gets 1$ to $m$} {
    $V \gets \emptyset$ \;
    \For {$i \gets 1$ to $n$}{
        \If{$p_{\text{ent}}(s_j, s^*_i) > \tau$ and $\phi(i, j) = 0$}{
            $V \gets V \cup i$\;
        }
    }
    $s^*_V \gets$ Concatenate sentences in $\{s^*_v \forall v \in V \}$\;
    \If{$V \text{ is consecutive and}$ $p_{\text{ent}}(s^*_V, s_j) > \tau$ }{
        \For {$v \in V$}{
            $\phi(v, j) \gets 1$\;
        }
    }
}

// Resolve 1-to-1 mappings\;
\For{$i \gets 1$ to $n$} {
    \For{$j \gets 1$ to $m$} {
        \If{$\phi(i, j) = 0$ and $p_{\text{ent}}(s_j, s^*_i) > \tau$ and $p_{\text{ent}}(s^*_i, s_j) > \tau$}{
            $\phi(i, j) \gets 1$\;
            
        }
    }
}
Return $\phi$\;

\end{algorithm}

\subsection{Entailment-induced Bipartite Graph}
\label{sec:bipartite_graph}
To determine which reference sentence has not been covered by the generated summary and which generated sentence is not faithful to the reference summary, we construct a bipartite graph that links sentences between a reference summary and a generated summary. An edge indicates the linked sentences contain equivalent information. If no edge connects to a reference sentence, we consider this sentence not covered by the generated summary. Similarly, if a generated sentence is not linked in the bipartite graph, this sentence is likely not faithful to the reference summary. We use the entailment probabilities computed by an NLI model to determine whether a pair of sentences contain equivalent information. The procedure of constructing the bipartite graph is shown in Algorithm \ref{alg:bipartite_graph}.

The NLI model takes in two sentences, a premise ($P$) and a hypothesis ($H$), and computes whether $P$ entails, contradicts, or is neutral to $H$. Here, we only focus on the entailment probability from the $i$-th reference sentence to the $j$-th generated sentence $p_{\text{ent}}(s^*_i, s_j)$. We use the \textsc{RoBERTa-Large} model\footnote{\url{https://huggingface.co/roberta-large-mnli}} trained on the MNLI dataset, achieving an accuracy of around 91\%, which is on par with the performance of state-of-the-art models. 

Let $\phi(i, j)$ denote the mapping between the $i$-th reference sentence and the $j$-th generated sentence. $\phi(i, j) = 1$ if a link exists between $s^*_i$ and $s_j$; otherwise, $\phi(i, j) = 0$. We first consider a simplified setting by assuming each reference sentence can be mapped to at most one generated sentence, and vice versa (i.e. $ 0 \leq \sum_j \phi(i, j) \leq 1$). In this setting, we can determine whether two sentences contain equivalent information by checking the entailment relation from both directions (lines 26-27). 

{
\small~
\begin{align}
    \phi(i, j)= 
\begin{cases}
    1,             & p_{\text{ent}}(s^*_i, s_j) > \tau \bigwedge p_{\text{ent}}(s_j, s^*_i) > \tau\\
    0,              & \text{otherwise}
\end{cases}
\end{align}
}
Here, $\tau$ is a hyperparameter that indicates the entailment threshold.

However, one reference sentence may contain information equivalent to multiple generated sentences (one-to-many mappings) and vice versa (many-to-one mappings). In \Cref{fig:mix-n-match}, for example, the second reference sentence contains information equivalent to the second and the third generated sentences combined. This relation cannot be discovered if we only check the entailment relation between pairs of individual sentences.

Therefore, we must resolve one-to-many and many-to-one mappings before checking one-to-one mappings. To find one-to-many mappings, for every reference sentence $s^*_i$, we look for consecutive generated sentences $\{s_j, s_{j+1}, ..., s_{j+k}\} ~~s.t.~ \max_{i}~ p_{\text{ent}}(s^*_i, s_m) > \tau ~~ \forall m \in \{j,...,j+k\}$ (lines 6-8). We only check for consecutive sentences based on our previous observation that dialogues are often summarized linearly. For every match, we concatenate the generated sentences $s_{j:j+k} = \{s_j, s_{j+1}, ..., s_{j+k}\}$ and check whether $s_{j:j+k}$ entails the reference sentence $s^*_{i}$ (lines 8-9). If the entailment holds, we let $\phi(i, m) = 1 ~~ \forall m \in \{j ,..., j+k\}$ (lines 11-12). The same approach is used to address many-to-one mappings (lines 14-22). Following Algorithm \ref{alg:bipartite_graph}, a bipartite graph is built between the generated summary and the reference summary. Henceforth, we denote the reference sentences that have not been covered as $\underline{S}^* = \{s^*_{i}|  \forall j ~~~ \phi(i, j) = 0  \}$ and generated sentences that can be mapped to some of the reference sentences as $\underline{S} = \{s_{j}|  \exists i ~~~ \phi(i, j) = 1 \}$.

\subsection{Uncovered Loss}
\label{sec:uncovered_loss}

The objective of the uncovered loss is to encourage the model to generate information from the reference summary that the generated summary has not covered. To this end, we train the model with \mixandmatch~ summaries, which are constructed by combining reference sentences that are not covered by the generated summary and generated sentences that contain information equivalent to some of the reference sentences. An example is shown in \Cref{fig:mix-n-match}.

The \mixandmatch summary $\hat{S}$ is constructed by taking the union of $\underline{S}$ and $\underline{S}^*$ and sorting the sentences by their index,
{
\small
\begin{align}
    \hat{S} = \textsc{Sort}(\underline{S} \cup \underline{S}^*). 
\end{align}
}

The uncovered loss is effectively maximum likelihood estimation (MLE) with \mixandmatch~ summaries being the decoding targets:

{
\small
\begin{align}
    \mathcal{L}_{\text{Uncovered}} = - \sum_{t}  \log p(\hat{S}_t| \hat{S}_{<t}, \mathcal{D}),
    \label{eq:uncovered_loss}
\end{align}
}
where $\mathcal{D}$ is the original dialogue and $\hat{S}_t$ denotes the $t$-th token in the \mixandmatch~ summary. 

The main advantages of constructing \mixandmatch~ summaries over other positive sample construction approaches, such as back translation and paraphrasing, are the two desired properties of this formulation. First, the model already has a high probability of generating sentences in $\underline{S}$. Therefore, the loss function (\Cref{eq:uncovered_loss}) does not penalize the model much for generating these sentences. Second, the penalty for generating sentences $\underline{S}^*$ is larger since the model has a lower probability of generating those sentences.

\begin{table*}[t]
    \small
    \centering
    \begin{adjustbox}{max width=\textwidth}
    {
    \begin{tabular}{lcccccccccccccccc}
        \toprule
        
        & \multicolumn{8}{c}{\textbf{\dialogsum~}} & 
        \multicolumn{8}{c}{\textbf{\samsum~}}  \\
        \cmidrule(lr){2-9} 
        \cmidrule(lr){10-17}
        \textbf{Model}      & $\text{RL}_{F}$ & $\text{RL}_{R}$ & $\text{BS}_{F}$ & $\text{BS}_{R}$ &  $\text{FC}_{F}$ & $\text{FC}_{R}$ & QS & QFE & $\text{RL}_{F}$ & $\text{RL}_{R}$ & $\text{BS}_{F}$ & $\text{BS}_{R}$ &  $\text{FC}_{F}$ & $\text{FC}_{R}$ & QS & QFE  \\ 
        \midrule
        TextRank~  & 27.74 & 29.16 & -3.000 & -3.039 & 60.55 & 59.54 & -1.948 & 0.566 & 15.08 & 16.15 & -4.374 & -3.891 & 34.28 & 33.02 & -2.172 & 0.237  \\

        \textsc{BART-Large} & 50.82 & 56.78 & -2.012
  & -1.960 & 82.90 & 85.86 & -1.183 & 1.854 & 49.53 & 52.71 & -2.248 & -2.332 &  62.46 & 61.28  & -0.912 & 2.335\\
        \ctrldiasumm~ & 48.99 & 57.25 & -2.145 & -1.985 & 82.55 & 85.96 & -1.214 & 1.817 & 47.79 & 51.17 & -2.360 & -2.414 & 61.50 & 61.76  & -0.957 & 2.272  \\
        
        \cods~  & 48.51 & 48.36 & -2.379 & -2.214 & 83.33 & 86.81 & -1.246 & 1.860 & 48.39 & 47.68 & -2.643 & -2.593 & 61.21 & 62.01  & -0.867 & 2.345 \\
        \conseq~  & 22.82 & 19.50 & -3.480 & -3.588 & 84.24 & 73.14 & -1.474 & 0.208 & 12.04 & 7.62 & -5.908 & -7.278 & 41.23 & 13.77 & -2.058 & 0.035\\
        \cliff~ & 51.87 & 56.22 & -2.012 & -1.973 & 85.38 & 86.30 & -1.106 & 2.109 & 43.70 & 45.49 & -2.485 & -2.340 & 55.47 & 56.01 & -1.063 & 1.891 \\ 
        \confit~  & 50.44 & 55.65 &   -2.049 & -2.016 
  & 83.34 & 86.37 & -1.179 & 1.790 & 49.29 & 52.76 & -2.188 & -2.316 & \textbf{65.03} & 63.12 & \textbf{-0.819} & 2.343  \\

        \cmidrule(lr){1-17}

        \modelshort~ & \textbf{51.96} & 59.04$^*$ & \textbf{-1.999}$^*$ & -1.904$^*$ & \textbf{86.48} & \textbf{89.03} & -1.082$^*$ & 2.087 & \textbf{50.08} & 52.91 & -2.228 & -2.310$^*$ & 64.19 & \textbf{63.52} & -0.829 & \textbf{2.407}$^*$\\ 
        ~~~ - $\mathcal{L}_{\text{Uncovered}}$ & 50.94 & \textbf{60.06}$^*$ & -2.044 & \textbf{-1.895}$^*$ & 83.26 & 87.45 & \textbf{-1.075}$^*$ & 2.339$^*$ & 49.78 & 53.57 &  -2.231 & -2.295$^*$ & 63.81 & 63.11 & -0.876 & 1.989   \\
        ~~~ - $\mathcal{L}_{\text{Contrastive}}$  & 51.53 & 59.27$^*$ & -2.012 & -1.901$^*$ & 82.90 & 85.86 & -1.130 & \textbf{2.399}$^*$ &  49.73 & \textbf{53.95} & \textbf{-2.185}$^*$ & \textbf{-2.143}$^*$ & 63.47 & 63.15 & -0.886 & 2.027  \\

        \bottomrule
    \end{tabular}
    }
    \end{adjustbox}
    \caption{Performance comparison on \dialogsum~ and \samsum~. - $\mathcal{L}_{\text{Uncovered}}$ and - $\mathcal{L}_{\text{Contrastive}}$ denote variants of \modelshort~ by ablating the corresponding loss. RL denotes \textsc{ROUGE-L} (\%), BS denotes \textsc{BARTScore}, FC denotes \textsc{FactCC} (\%), QS denotes \textsc{QUALS}, and QFE denotes \textsc{QAFactEval}. The subscripts $F$ and $R$ denote F1 score and recall, respectively. The proposed method outperforms previous systems on both \dialogsum~ and \samsum~ in most metrics, especially on the recall measures. Statistical significance over previous best systems computed with the permutation test \cite{fisher1937design} is indicated with * ($p < .01$). }%
    \label{tab:main}
\end{table*}
\subsection{Contrastive Loss}
\label{sec:contrastive_loss}
In the early stage of our experiment, the original goal was to discourage the model from generating factually inconsistent sentences. We adopt unlikelihood training \cite{Welleck2020Neural} to decrease the probability of sampling these sentences from the model. However, we found that this objective causes the model to generate nonsense sequences. This phenomenon was also observed when we experimented with \conseq~ \cite{nan-etal-2021-improving}, which also incorporates such a loss function into its training process, as shown in \Cref{subsec:main_results}. We hypothesize that it resulted from the fact that sentences in dialogue summaries share similar structures. Hence, using the unlikelihood training objective would confuse the model. 

Instead, we pivoted our focus on differentiating factually consistent sentences from their inconsistent counterparts with the proposed contrastive loss. For each summary, we use the factually inconsistent sentences as negative samples (i.e. $s_j \notin \underline{S}$) and consistent sentences as positive samples (i.e. $s_j \in \underline{S}$). The contrastive learning objective takes a similar form as the InfoNCE loss \cite{oord2018representation}: \looseness=-1

{
\small
\begin{align}
    \mathcal{L}_{\text{Contrastive}} = - \sum_{s_i \in \underline{S}} \frac{\exp(cos(h_i, h_{S^*}))}{\sum_{s_j \in {S}} \exp(cos(h_j, h_{S^*}))}
    \label{eq:contrastive}
\end{align}
}
, where $h_i$ and $h_j$ denote the representations of the generated sentences, $h_{S^*}$ means the representations of the reference summary, and $cos(,)$ denotes cosine similarity.
The main difference between our contrastive objective and the other work \cite{cao-wang-2021-cliff, tang-et-al-2022-confit} is granularity. \Cref{eq:contrastive} operates at the sentence level rather than the summary level; therefore, it provides finer-grained training signals.\linepenalty=1000 %

\subsection{Training}
The final loss function that our model is optimized with is a weighted sum of the two aforementioned loss functions and MLE,

{
\vspace{-2mm}
\small
\begin{align}
    \mathcal{L}_{\text{Final}} =  \mathcal{L}_{\text{MLE}} + \alpha \mathcal{L}_{\text{Uncovered}} + \beta  \mathcal{L}_{\text{Contrastive}},
    \label{eq:final_loss}
\end{align}
}
where $\mathcal{L}_{\text{MLE}}$ is:

{
\vspace{-2mm}
\small
\begin{align}
    \mathcal{L}_{\text{MLE}} = - \sum_{t}  \log p(S^*_t| S^*_{<t}, \mathcal{D}).
    \label{eq:mle_loss}
\end{align}
\vspace{-2mm}
}

\section{Experiments}

\subsection{Datasets}
Experiments are conducted on two English-language dialogue summarization datasets: \samsum~ \cite{gliwa-etal-2019-samsum} and \dialogsum~ \cite{chen-etal-2021-dialogsum}. \samsum~ contains 16,369 online chitchat dialogues with an average of around 94 tokens per dialogue. \dialogsum~ is a spoken dialogue dataset that consists of 13,460 samples in total. With an average token count of about 131, the dialogues in \dialogsum~ are under real-life scenarios with clear communication patterns and intents. Details of the dataset statistics can be found in \Cref{sec:dataset_stats}.\looseness=-1

\subsection{Metrics}
Our evaluation focuses on measuring the factual consistency, particularly the missing information challenge, of the summarization models. Therefore, we adopt recently developed metrics that have been shown to correlate well with human judgments in terms of faithfulness. BARTScore \cite{yuan2021bartscore} computes the semantic overlap between the generated summary and the reference summary by calculating the logarithmic probability of generating each summary conditioned on the other one. Since our goal is to assess how well the model reduce information missing from the reference summary, we consider the \textit{Recall (R)} setting where we assess $p(S^*|S, \theta)$, the likelihood of generating the reference summary $S$ given the generated summary $S^*$. FactCC \cite{kryscinski-etal-2020-evaluating} is an entailment-based metric that predicts the faithfulness probability of a claim w.r.t. with the source texts. Similar to BARTScore, we use FactCC in the \textit{Recall} setting where the claim is a reference sentence and the source text is the generated summary. We report the mean of the average \textsc{Correct} probability of each sentence within a generated summary. %

In addition, we report the ROUGE-L metric \cite{lin-2004-rouge}, which has been also shown to better reflect faithfulness compared to ROUGE-1 and ROUGE-2 \cite{pagnoni-etal-2021-understanding}. For these metrics, we also consider the \textit{F1} setting, where we compute each metric in the reverse direction ($S^* \rightarrow S$) and then take the average of both directions, to validate that the model is not generating too much redundant information. Finally, two recently introduced QA-based metrics that have demonstrated close approximation to human judgements in terms of factuality, \textsc{QUALS} \cite{nan-etal-2021-improving} and \textsc{QAFactEval} \cite{fabbri-etal-2022-qafacteval}, are also used for evaluation.

\begin{figure*}[t]
    \centering
    \begin{subfigure}[t]{0.48\textwidth}

    \includegraphics[width=.95\linewidth]{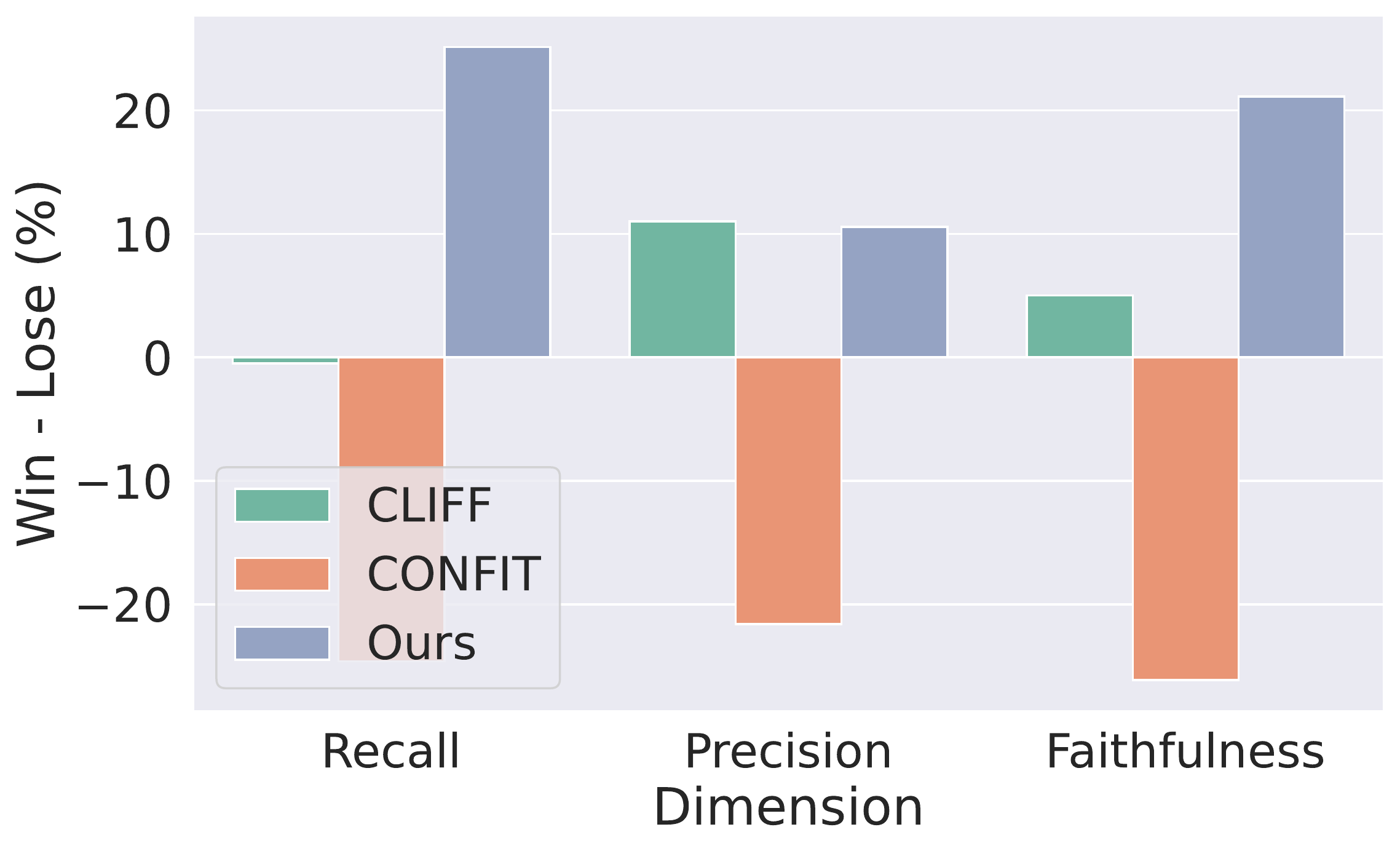}
    \caption{Human evaluation results on \dialogsum~.}
    \end{subfigure}
    ~
    \begin{subfigure}[t]{0.48\textwidth}
    \includegraphics[width=.95\linewidth]{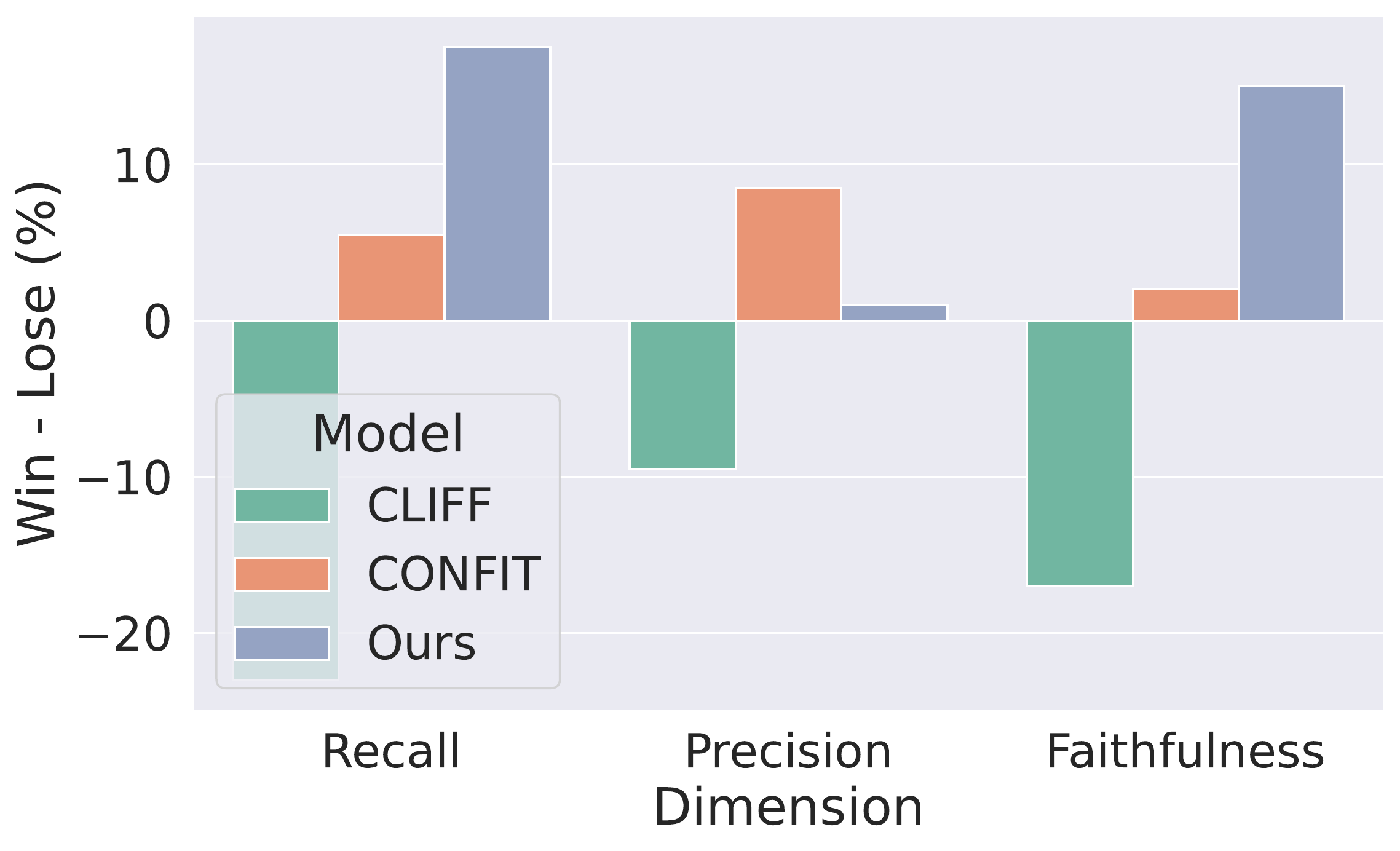}
    \caption{Human evaluation results on \samsum~.}
    \end{subfigure}

    \caption{Human evaluation results. \modelshort~ achieves the highest \textsc{Recall} and \textsc{Faithfulness} scores on both datasets, suggesting the advantages of our approach in reducing missing information and improving the overall faithfulness of the generated dialogue summary.}
    \label{fig:human_eval}
    \vspace{-5mm}
\end{figure*}

\subsection{Implementation Details}
We choose \textsc{BART} \cite{lewis-etal-2020-bart} as the backbone seq2seq model as it has demonstrated better dialogue summarization performance than other pre-trained language models \cite{tang-et-al-2022-confit}, such as PEGASUS \cite{zhang-2020-pegasus} and T5 \cite{2020t5}. The proposed models are optimized using AdamW \cite{loshchilov2018decoupled} with learning rate 3e-5 and weight decay 1e-3. The maximum input sequence length is set to 1024. For all baseline models, we use the best hyper-parameters reported in their papers. We fix $\tau$ to be 0.5 throughout all our experiments. $\alpha$ and $\beta$ are both 1.0. \looseness=-1

\subsection{Baselines}
We compare \modelshort~ with the following competitive baseline systems. \textbf{TextRank} \cite{mihalcea-tarau-2004-textrank} is a graph-based ranking algorithm that performs extractive summarization. 
\textbf{\textsc{BART}} \cite{lewis-etal-2020-bart} is a seq2seq language model pre-trained on various denoising objectives. \textbf{\ctrldiasumm~} \cite{liu-chen-2021-controllable} and \textbf{\cods~} \cite{wu-etal-2021-controllable} are controllable generation frameworks that generate summaries guided by named entity planning and sketches, respectively. \textbf{\conseq~} \cite{nan-etal-2021-improving} learns a contrastive objective based on unlikelihood training, where positive and negative samples are selected by \textsc{QUALS}. \textbf{\cliff~} \cite{cao-wang-2021-cliff} and \textbf{\confit~} \cite{tang-et-al-2022-confit} are trained with a similar contrastive learning loss that takes the form of the InfoNCE loss \cite{oord2018representation}, except that \confit~ is optimized with an additional self-supervised loss that aims to reduce reference errors. \textsc{BART-Large} is used across all experiments that involve pre-trained language models for fair comparison.

\section{Results}

\begin{table*}[t]
    \small
    \centering
    {
    \begin{tabular}{p{0.33\linewidth} p{0.3\linewidth}p{0.3\linewidth}}
        \toprule
        
        \textbf{Reference Summary} & \textbf{\confit~} & \textbf{\modelshort~} \\
        
        \midrule
        Mike took his car into garage today. Ernest is relieved as someone had just crashed into a red Honda which looks like Mike's. & Mike took his car to the garage today. Someone crashed into his car. & Mike took his car into the garage today. Someone just crashed into \hlc{lightblue}{a red Honda looking like Mike's}. \\
        \midrule
        Hilary has the keys to the apartment. Benjamin wants to get them and go take a nap. Hilary is having lunch with some French people at La Cantina. Hilary is meeting them at the entrance to the conference hall at 2 pm. Benjamin and Elliot might join them. They're meeting for the drinks in the evening. & Benjamin, Elliot, Daniel and Hilary will meet at La Cantina at 2 pm to have lunch with some French people who work on the history of food in colonial Mexico. They will try to avoid talking about their subject of research. &  \hlc{lightblue}{Hilary has the keys to Benjamin, Elliot and Daniel's apartment.} They will meet at the entrance to the conference hall at 2 pm and go to La Cantina for lunch with some French people who work on the history of food in colonial Mexico.\\
        \bottomrule
    \end{tabular}
    }
    \vspace{-2mm}
    \caption{Qualitative analysis on the outputs of \modelshort~ and \confit~. The two rows demonstrate the \textit{missing details} and the \textit{missing sentences} issue of the summaries generated by \confit~, respectively. The extra information in the outputs of \confit~ that also occurs in the reference summaries is \hlc{lightblue}{highlighted in blue}. In both cases, \modelshort~ is able to cover more content presented in the reference summaries.}
    \label{tab:qualitative_analysis}
    
\end{table*}

\subsection{Main results}
\label{subsec:main_results}
\Cref{tab:main} summarizes the main results on \dialogsum~ and \samsum~. \modelshort~ outperforms previous approaches in almost all metrics, especially recall measures. This result reflects that the proposed approach generates summaries that cover more content in the reference summaries lexically and semantically. One interesting observation was the deficient performance of \conseq~ on both datasets. We hypothesize that poor performance was the use of the unlikelihood training objective in their loss, as mentioned in \Cref{sec:contrastive_loss}. Since sentences of dialogue summaries often share similar structures, adopting such an objective could confuse the model. We verified this hypothesis by running a small experiment by training \textsc{BART-Large} with MLE and negative samples determined by \textsc{QUALS}, similar to \conseq~. The resulting model also produces significantly lower performance than training with MLE alone. The finding confirms that the poor performance of \conseq~ is caused by the unlikelihood training and that such a loss function is unsuitable for dialogue summarization. %

\subsection{Human Evaluation}
\label{subsec:human_eval}

To further validate the effectiveness of \modelshort~, we use Amazon's Mechanical Turk (AMT) to recruit workers to conduct human evaluations on three methods: \cliff~, \confit, and \modelshort~. We sampled 100 dialogues from the test set of \dialogsum~ and \samsum~, respectively. For each dialogue, human judges are presented with a pair of summaries produced by two different approaches and asked to select the better one with respect to three dimensions. \textbf{\textsc{Recall}} assesses the portion of information in the reference summary covered by the generated summary. \textbf{\textsc{Precision}} considers whether all the content in the generated summary occurs in the reference summary. \textbf{\textsc{Faithfulness}} examines whether the generated summary is factually consistent with the dialogue. "Tie" is selected if the judges consider the two summaries to be of equal quality. The final score of each system is calculated as the percentage of times the system is selected as the better one minus the percentage of times the system is not. To evaluate the annotation quality, we compute the inter-annotator agreement. The average Cohan's Kappa \cite{cohen1960coefficient} is 54.35\%, indicating a moderate agreement. Details of the human evaluation setup can be found in \Cref{apx:eval_details}. 

The human evaluation results are demonstrated in \Cref{fig:human_eval}. We have the following observations. First, \modelshort~ achieves the highest \textsc{Recall} scores on both datasets, indicating that our approach is the best in addressing the missing information issue for dialogue summarization. Second, while \modelshort~ does not score the highest on \textsc{Precision}, we achieve the highest scores on \textsc{Faithfulness}. This implies that even though our approach often generates summaries with extra information, the additional content is likely still faithful to the input. To measure the amount of additional information produced, we compute the average number of tokens per summary for each model. As seen in \Cref{tab:summary_stats}, the summaries generated by \modelshort~ is only slightly longer than those produced by \cliff~ and \confit~. This suggests that \modelshort~ achieves significantly higher faithfulness and coverage than \cliff~ and \confit~ while maintaining conciseness. \looseness=-1

\begin{table}[h]
    \small
    \centering
    \begin{adjustbox}{max width=\textwidth/2}
    {
    \begin{tabular}{lcc}
        \toprule
        
        Model & \dialogsum~ & \samsum~ \\
        \midrule
        \textsc{ConFit} & 29.46 & 22.45   \\
        \textsc{Cliff} & 27.34  &  22.30  \\
        \textsc{BART-Large} & 28.03 & 23.19   \\
        \midrule
        \modelshort~ & 31.32 & 24.23  \\
        \bottomrule
    \end{tabular}
    }
    \end{adjustbox}
    
    \caption{Average token count per summary generated by different models.}
    \label{tab:summary_stats}
    
\end{table}

\subsection{Qualitative Analysis}

To provide better insight into the effectiveness of the proposed method, we conduct a qualitative analysis using the 100 dialogues randomly sampled from the \samsum~ dataset. Specifically, we further categorize missing information errors into two sub-types: (1) \textbf{\textit{missing details}} where partial information of a sentence in the reference summary is missing in the generated summary and (2) \textbf{\textit{missing sentences}} where the model fails to generate an entire sentence in the reference summary. An example of each sub-type is shown in \Cref{tab:qualitative_analysis}. By comparing the test sets outputs of \confit~ and \modelshort~, we see that there are 10 improved cases with less \textit{missing details} and 6 cases where \textit{missing sentences} is mitigated by \modelshort~. Meanwhile, our proposed approach only introduces \textit{missing details} error and \textit{missing sentences} error in 1 and 2 examples, respectively. This implies that our approach is effective in alleviating both sub-types of missing information error while particularly advantageous in reducing \textit{missing details} errors.

\subsection{Correlation with Human Judgements}
Although recently proposed metrics have been shown to be highly correlated with human judgments on news summarization in terms of factuality \cite{kryscinski-etal-2020-evaluating, yuan2021bartscore}, no previous work has studied the transferability of these metrics to dialogue summarization. We seek to answer this question by computing the correlation of the automatic metrics in \Cref{tab:main} with the human annotations discussed in \Cref{subsec:human_eval}. Using Kendall's Tau \cite{kendall1938new} as the correlation measure, the results are summarized in \Cref{tab:metric_correlation}. We observe that: (1) $\textsc{BARTScore}_{R}$ is the most consistent and reliable metric across the three dimensions. It performs the best in \textsc{Recall} on both datasets, indicating that $\textsc{BARTScore}_{R}$ is most suitable for measuring how well a model resolves the missing information issue in dialogue summarization. (2) Although a large number of invalid questions and answers are generated, \textsc{QUALS} is the best metric for assessing \textsc{Precision} overall. (3) $\textsc{FactCC}_{F}$ and $\textsc{FactCC}_{R}$ are two of the worst metrics in general. This could be explained by the fact that \textsc{FactCC} constructs negative samples with some semantically variant transformations. However, these transformations may not be comprehensive enough to cover all cases. Hence, the poor transferability of \textsc{FactCC} on these two datasets. %

\begin{table*}[h]
    \small
    \centering
    \begin{adjustbox}{max width=\textwidth}
    {
    \begin{tabular}{lcccccc}
        \toprule
        
         & \multicolumn{3}{c}{\textbf{\dialogsum~}} & 
        \multicolumn{3}{c}{\textbf{\samsum~}}  \\
        \cmidrule(lr){2-4} 
        \cmidrule(lr){5-7}
        Metric & \textsc{Recall} & \textsc{Precision} & \textsc{Faithfulness} & \textsc{Recall} & \textsc{Precision} & \textsc{Faithfulness} \\

        \midrule
        $\textsc{ROUGE-L}_{F}$  & 23.50 & \textbf{24.21} & 10.29 & 6.07 & 10.24 & -0.75  \\
        $\textsc{ROUGE-L}_{R}$ &  23.46 & 2.51 & 4.24 & 29.52 & 9.61 & 17.88\\
        $\textsc{BARTScore}_{F}$ & 18.35 & 25.94 & 3.17 & 15.50 & 8.00 & 10.69 \\
        $\textsc{BARTScore}_{R}$ & \textbf{26.48} & 14.87 & 9.25 & \textbf{32.10} & 9.68 & \textbf{24.11}\\
        $\textsc{FactCC}_{F}$  & 6.15 & 6.93 & 1.19  &-3.43 & 5.12 & -2.28 \\
        $\textsc{FactCC}_{R}$ & 4.79 & 6.86 & 10.56&  4.13 & 10.32 &-1.43 \\
        $\textsc{QUALS}$ & 14.23 & 23.61 & -0.83 & 1.55 & \textbf{15.35} & 4.50 \\
        $\textsc{QAFactEval}$ & 14.06 & 16.20 & \textbf{16.80} & 5.03 & 2.83 & 6.26\\
        \bottomrule
    \end{tabular}
    }
    \end{adjustbox}
    \caption{Correlation (\%) of automatic metrics with human judgements. We first convert human evaluation results and automatic metric scores into a scale of \{-1, 0, 1\}, which corresponds to \{\textsc{Lose}, \textsc{Tie}, \textsc{Win}\}. Then, Kendall's Tau \cite{kendall1938new} is used to compute the correlation between two sequences.}
    \label{tab:metric_correlation}
\end{table*}

\begin{figure}[h]
    \centering
    \includegraphics[width=0.9\linewidth]{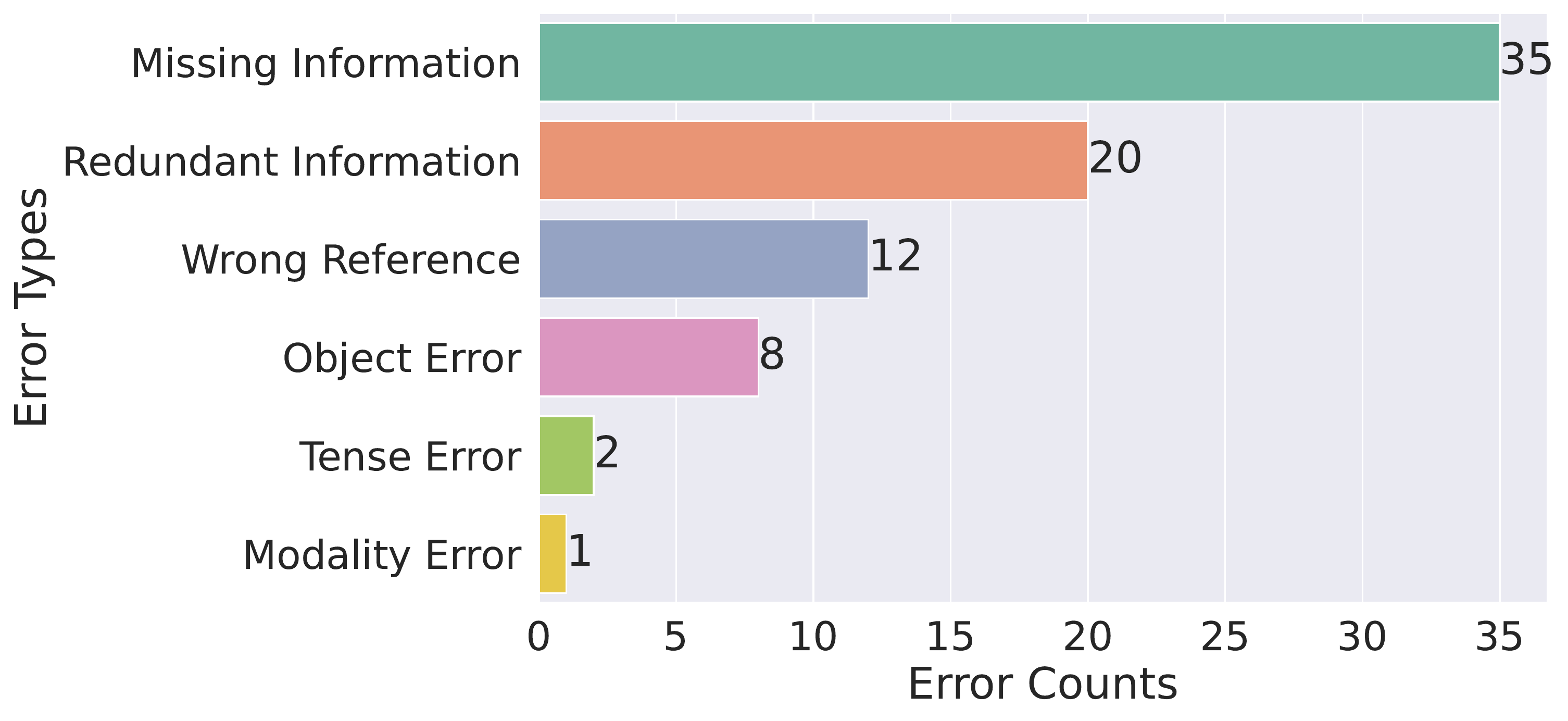}
    \caption{Remaining challenges.}
    
    \label{fig:remaining_challenges}
\end{figure}
\subsection{Remaining Challenges}

We analyzed the remaining errors by comparing 100 generated summaries with corresponding reference summaries on the \samsum~ datasets using the categories of factual errors defined in \citet{tang-et-al-2022-confit}. The results are shown in \Cref{fig:remaining_challenges}. We observe that missing information still accounts for the largest portion of factual errors, even though our approach significantly exceeds prior methods in mitigating this issue. This reflects that this issue is challenging to tackle and that there is still a great opportunity to improve the reduction of missing information. As a comparison, we manually inspected outputs of \textsc{BART-Large} using the same 100 dialogues as input. We found 42 cases where information is missing from the dialogue summaries produced by \textsc{BART-Large}. This observation further confirms the effectiveness of \modelshort~ in addressing insufficient coverage.  In addition, redundant information is another major source of errors. Although we have shown in \Cref{subsec:human_eval} that the additional information generated by \modelshort~ is likely still faithful to the input dialogue, compactness is one of the important qualities of a summary. This can be improved by using NLI to guide the model to avoid generating extra information. Other common mistakes are wrong reference and object errors, both of which can be addressed with the self-supervised loss discussed in \citet{tang-et-al-2022-confit}.\footnote{This analysis is not comparable to results reported in \citet{tang-et-al-2022-confit} due to differences in the sampled examples.}

\section{Related Work}

\paragraph{Dialogue Summarization} Early work on dialogue summarization focus on the AMI meeting corpus \cite{McCowan2005ami} due to the lack of dialogue summarization data. These studies enhance summarization performance by leveraging features of conversational data, such as dialogue act \cite{DBLP:conf/slt/GooC18}, visual features \cite{li-etal-2019-keep}, and the relationships between summary and dialogue \cite{oya-etal-2014-template}. Later, \citet{gliwa-etal-2019-samsum} released the \samsum~ dataset, the first large-scale dialogue summarization dataset, enabling abstractive summarization research on casual chat dialogue. With the rise of large language models (LMs), recent work focuses on improving the controllability of sequence-to-sequence models built upon large LMs. For instance, \citet{wu-etal-2021-controllable} proposes to utilize a summary sketch to control the granularity of the summary generated. \citet{liu-chen-2021-controllable} conditions the generators with person name entities to control which people to include in the generating summary. \citet{chan-etal-2021-controllable} improves controllability by formulating the summarization task as a constrained Markov Decision Process. \looseness=-1

\paragraph{Factual Consistency Enhancement}
While factuality has been widely explored in the field of fact-checking and fake news detection \cite{thorne-etal-2018-fever, wadden-etal-2020-fact, huang-etal-2022-concrete, shu2018fakenewsnet, pan-etal-2021-zero, huang2022faking}, factual inconsistency remains a major challenge for abstractive summarization. One line of work attempts to improve the faithfulness of the generated summary with a separate correction model that corrects the errors made by the summarization model \cite{dong-etal-2020-multi, cao-etal-2020-factual, fabbri2022improving} or directly fix factual inconsistencies in the training data \cite{adams2022learning}. Another line of work employs auxiliary loss functions to improve models' representations or discourage the model from generating unfaithful outputs \cite{cao-wang-2021-cliff, DBLP:journals/kbs/ChenLCK21, nan-etal-2021-improving, tang-et-al-2022-confit}. The main advantage of these approaches is the efficiency in inference time. \looseness=-1 %

Some studies have attempted to use NLI to detect factual inconsistency in generated summaries. Early approaches rely on out-of-the-box NLI models, which did not yield satisfactory results \cite{falke-etal-2019-ranking}. \citet{Barrantes2020AdversarialNF} improved the detection accuracy by using an NLI model fine-tuned on the Adversarial NLI dataset \cite{nie-etal-2020-adversarial}. \citet{laban-etal-2022-summac} addresses the mismatch issue in input granularity between NLI datasets and inconsistency detection by passing sentence pairs as inputs instead of document-summary pairs. \citet{kryscinski-etal-2020-evaluating} and \citet{yin-etal-2021-docnli} trains document-sentence entailment models to address the granularity mismatch issue. \citet{utama-etal-2022-falsesum} introduces a controllable generation framework that generates document-level NLI training data for identifying factual inconsistency. Our work leverages an NLI model to guide the dialogue summarization model to recover missing information. 

\section{Conclusion}
We have proposed \modelshort~, a dialogue summarization framework that generates summaries with mitigated missing information and improved faithfulness. To instruct the model to generate missing content from the reference summaries and to differentiate factually consistent generated sentences from their inconsistent counterparts, we propose two losses based on NLI. Experimental results on the \dialogsum~ and \samsum~ datasets showed that our approach achieves significantly higher faithfulness and coverage, while still maintaining conciseness, compared to prior methods. In addition, we measure the correlation between the reported automatic metrics and human judgments to provide insight into the most suitable metric for evaluating the coverage and factuality of dialogue summaries for future research.%

\section{Ethical Considerations}

We acknowledge that the use of large language models pre-trained on the Web could lead to biased outputs. We did find out that our model may sometimes generate the incorrect pronouns for neutral names. For example, in \Cref{fig:toy_example}, Charlee is being referred to as a male in the generated summary, while Charlee is actually a female as shown in the reference summary. Such an issue is often caused by under-specified context (e.g. Charlee's gender is not mentioned in the input dialogue). Fortunately, we found that such an error accounts for < 1\% of the total outputs from our framework and the issue can be largely alleviated when enough context is provided. \looseness=-1

\section{Limitations}
While our proposed approach is effective in mitigating missing information, this issue is still far from resolved, as shown in \Cref{fig:remaining_challenges}. Significant effort is needed to ensure dialogue summarization models produce completely factual content. In addition, our method works as we found that most of the reference summaries in the two datasets we used are faithful to the corresponding dialogue. The proposed method may not work on other summarization datasets, such as XSum, which contains hallucinations in about 70\% of the reference summaries \cite{maynez-etal-2020-faithfulness}.
\section{Acknowledgments}
We would like to extend our gratitude to the reviewers for their valuable feedback and insights, which greatly contributed to the improvement of this paper. We would also like to thank the human evaluators for their time and effort in assessing the performance of our model. Their contributions have been essential in ensuring the quality of our research.

\bibliography{anthology,custom}
\bibliographystyle{acl_natbib}
\clearpage
\newpage
\appendix

\section{Dataset Statistics}
We present the detailed statistics of \dialogsum~ and \samsum~ in \Cref{tab:dataset_stats}.

\begin{table}[h]
    \small
    \centering
    \begin{adjustbox}{max width=\textwidth/2}
    {
    \begin{tabular}{cccc}
        \toprule
        
        Dataset & \# Dialogues & Avg. Dialogue Words & Avg. Summ. Words\\
        \midrule
        \dialogsum~ & 13,460 & 187.5 & 31.0  \\
        \samsum~ & 16,369 & 124.1 & 23.4 \\
        \bottomrule
    \end{tabular}
    }
    \end{adjustbox}
    
    \caption{Statistics of \dialogsum~ and \samsum~. We use the NLTK tokenizer to compute word counts for both datasets.}
    \label{tab:dataset_stats}
    
\end{table}
\label{sec:dataset_stats}

\section{Human Evaluation Details}
\label{apx:eval_details}
\begin{figure*}[bt]
    \centering
    \includegraphics[width=0.9\linewidth]{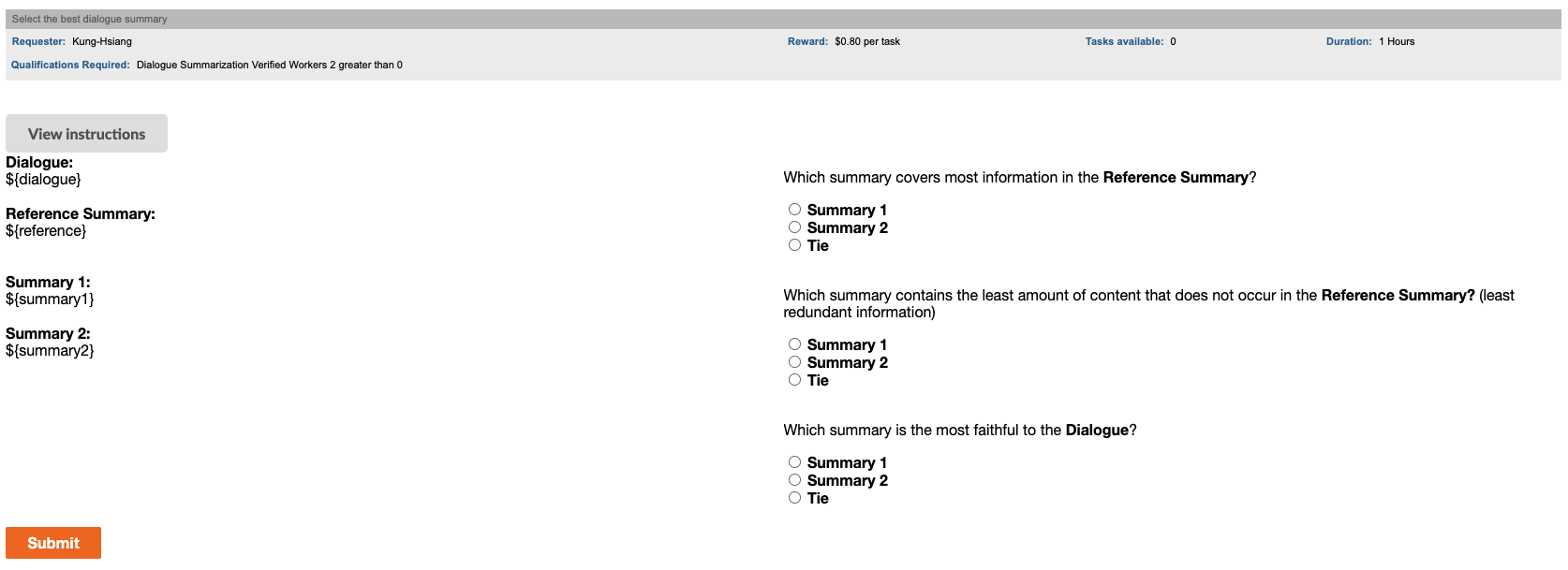}
    \caption{MTurk UI for our human evaluation.}
    \label{fig:mturk_ui}
\end{figure*}

In this section, we describe the details of our human evaluation. We recruit AMT workers from the United States for ensuring language fluency. Qualification requirements are set such that only workers who have an acceptance rate greater than 99\% and have more than 10,000 accepted HITs in the past are allowed to work on our annotation task. To further ensure annotation quality, we conducted two rounds of annotations. In the first round, we launched 100 HITs to select high-quality annotators in the first round. 8 qualified annotators are selected to enter the second round to conduct the remaining evaluation. We set the reward to \$0.8 per HIT to encourage experienced annotators to participate. Our annotation interface is displayed in \Cref{fig:mturk_ui}.

For each HIT, annotators are provided with a piece of dialogue and a corresponding reference summary as well as two summaries generated from different systems, demonstrated on the left segment of the interface. Based on the summaries and the dialogue, annotators are tasked to answer three questions shown on the right segment of the interface, each of which corresponds to \textsc{Recall}, \textsc{Precision}, and \textsc{Faithfulness}. They need to determine which summary is better with regard to each prompt. 

\section{Comparison with Other Data Augmentation Methods}
We compared our \mixandmatch~ summary construction technique with other data augmentation methods, including back translation (\textsc{BackTranslate}) and paraphrasing (\textsc{Paraphrasing}). For back translation, we use mBART-50 \cite{tang2020multilingual} to translate a summary from English to German and then back to English. For paraphrase generation, we use this open source package\footnote{\url{https://github.com/Vamsi995/Paraphrase-Generator}}. The experimental results are summarized in \Cref{tab:other_da}. Training with \mixandmatch~ summaries achieves the highest scores on most metrics, indicating that our proposed method is the most effective in improving the factuality of the generated summaries.
\begin{table*}[t]
    \small
    \centering
    \begin{adjustbox}{max width=\textwidth}
    {
    \begin{tabular}{lcccccccccccccccc}
        \toprule
        
        & \multicolumn{8}{c}{\textbf{\dialogsum~}} & 
        \multicolumn{8}{c}{\textbf{\samsum~}}  \\
        \cmidrule(lr){2-9} 
        \cmidrule(lr){10-17}
        \textbf{Model}      & $\text{RL}_{F}$ & $\text{RL}_{R}$ & $\text{BS}_{F}$ & $\text{BS}_{R}$ &  $\text{FC}_{F}$ & $\text{FC}_{R}$ & QS & QFE & $\text{RL}_{F}$ & $\text{RL}_{R}$ & $\text{BS}_{F}$ & $\text{BS}_{R}$ &  $\text{FC}_{F}$ & $\text{FC}_{R}$ & QS & QFE  \\ 
        \midrule

        \mixandmatch~ & \textbf{51.53} & \textbf{59.27} & \textbf{-2.012} & \textbf{-1.901} & 82.90 & 85.86 & \textbf{-1.130} & \textbf{2.399} &  \textbf{49.73} & \textbf{53.95} & \textbf{-2.185} & \textbf{-2.143} & 63.47 & 63.15 & \textbf{-0.886} & \textbf{2.027}\\ 
        \textsc{BackTranslate} &  50.41  & 58.22 & \textbf{-2.012} & -2.032 & \textbf{83.20} & 84.23 & -1.230 & 2.245 & 49.02 & 52.93 & -2.234 & -2.159 & 64.69 & 62.10 & -1.230 & 1.984\\
        \textsc{Paraphrasing}  &  50.32 & 59.22 & -2.133 & -1.936 & 82.20 & \textbf{87.62} & -1.198 & 2.333 & 49.23  & 53.94 & -2.320 & -2.178 & \textbf{64.78} & \textbf{63.98}  & -1.130 & 2.015\\

        \bottomrule
    \end{tabular}
    }
    \end{adjustbox}
    \vspace{-2mm}
    \caption{Performance comparison on \dialogsum~ and \samsum~ with other positive data augmentation methods. }
    \label{tab:other_da}
    \vspace{-5mm}
\end{table*}

\section{Hardware and Software configurations}
All experiments are conducted on a Linux machine with NVIDIA V100. We use PyTorch 1.11.0 with CUDA 10.1 as the Deep Learning framework and utilize Transformers 4.19.2 to load all pre-trained language models.

\section{Validation Set Performance}

We report the validation set performance of our proposed model in \Cref{tab:val_performance}.

\begin{table*}[t]
    \small
    \centering
    \begin{adjustbox}{max width=\textwidth}
    {
    \begin{tabular}{lcccccccccccc}
        \toprule
        
        & \multicolumn{6}{c}{\textbf{\dialogsum~}} & 
        \multicolumn{6}{c}{\textbf{\samsum~}}  \\
        \cmidrule(lr){2-7} 
        \cmidrule(lr){8-13}
        \textbf{Model}      & $\text{RL}_{F}$ & $\text{RL}_{R}$ & $\text{BS}_{F}$ & $\text{BS}_{R}$ &  $\text{FC}_{F}$ & $\text{FC}_{R}$ & $\text{RL}_{F}$ & $\text{RL}_{R}$ & $\text{BS}_{F}$ & $\text{BS}_{R}$ &  $\text{FC}_{F}$ & $\text{FC}_{R}$   \\ 
        \midrule

        \modelshort~ & 48.45 & 51.27 & -2.149 & -2.169 & 71.36 & 70.65 & 50.61 & 53.74 & -2.212 & -2.134 & 64.27 & 64.56  \\

        \bottomrule
    \end{tabular}
    }
    \end{adjustbox}
    \vspace{-2mm}
    \caption{Validation set performance. }
    \label{tab:val_performance}
    \vspace{-5mm}
\end{table*}

\section{Number of Parameters}
We do not introduce additional parameters to the backbone language model, \textsc{BART-Large}. During training time, the number of parameters equals to the sum of the number of parameters in \textsc{BART-Large} and \textsc{RoBERTa-Large}. In inference time, since we do not need the NLI component, the number of parameters is the same as that of \textsc{BART-Large}.

\section{Scientific Artifacts}
The licenses for all the models and software used in this paper are listed below in parentheses: BART (MIT License), \textsc{FactCC} (BSD-3-Clause License), \textsc{QAFactEval} (BSD-3-Clause License), \textsc{BARTScore} (Apache License 2.0), \textsc{QUALS} (MIT License), py-ROUGE (Apache License 2.0), NLTK (Apache License 2.0).

\end{document}